\documentclass{article}

\PassOptionsToPackage{numbers}{natbib}
\usepackage[preprint]{neurips_2019}

\usepackage[utf8]{inputenc} 
\usepackage[T1]{fontenc}    
\usepackage{hyperref}       
\usepackage{cleveref}
\usepackage{url}            
\usepackage{booktabs}       
\usepackage{amsfonts}       
\usepackage{nicefrac}       
\usepackage{microtype}      
\usepackage{graphicx}
\usepackage{caption}
\usepackage{subcaption}
\usepackage{bm}
\usepackage{amsmath} 
\usepackage{amssymb}
\usepackage{amsthm}
\usepackage{amsxtra}
\usepackage{soul}               
\usepackage[color=green!40]{todonotes}

\title{Hierarchical Auxiliary Learning}

\author{%
  Jaehoon Cha \\
  \texttt{jaehoon.cha@xjtlu.edu.cn} \\
 \And
 Kyeong Soo Kim \\
  \texttt{kyeongsoo.kim@xjtlu.edu.cn} \\
 \And
 Sanghyuk Lee \\
  \texttt{sanghyuk.lee@xjtlu.edu.cn} \\
\AND
\\
 Department of Electrical and Electronic Engineering \\
 Xi'an Jiaotong-Liverpool University \\
 Suzhou, 215123, P. R. China \\
}
\graphicspath{{./plots/}, {./figures/}}

\begin{document}

\maketitle

\begin{abstract}
  Conventional application of convolutional neural networks (CNNs) for image
  classification and recognition is based on the assumption that all target
  classes are equal (i.e., no hierarchy) and exclusive of one another (i.e., no
  overlap). CNN-based image classifiers built on this assumption, therefore,
  cannot take into account an innate hierarchy among target classes (e.g., cats
  and dogs in animal image classification) or additional information that can be
  easily derived from the data (e.g., numbers larger than five in the
  recognition of handwritten digits), thereby resulting in scalability issues
  when the number of target classes is large. Combining two related but slightly
  different ideas of \textit{hierarchical classification} and \textit{logical
    learning by auxiliary inputs}, we propose a new learning framework called
  \textit{hierarchical auxiliary learning}, which not only address the
  scalability issues with a large number of classes but also could further
  reduce the classification/recognition errors with a reasonable number of
  classes. In the hierarchical auxiliary learning, target classes are
  semantically or non-semantically grouped into superclasses, which turns the
  original problem of mapping between an image and its target class into a new
  problem of mapping between a pair of an image and its superclass and the
  target class. To take the advantage of superclasses, we introduce an auxiliary
  block into a neural network, which generates auxiliary scores used as
  additional information for final classification/recognition; in this paper, we
  add the auxiliary block between the last residual block and the
  fully-connected output layer of the ResNet.
  Experimental results demonstrate that the proposed hierarchical auxiliary
  learning can reduce classification errors up to 0.56, 1.6 and 3.56 percent
  with MNIST, SVHN and CIFAR-10 datasets, respectively.
\end{abstract}

\section{Introduction}
Deep Convolutional Neural Networks (CNNs) have attracted a considerable
attention due to their superior performance in image classification
\cite{zagoruyko2016wide, szegedy2017inception, huang2017densely,
  gastaldi2017shake}. With residual blocks, the depth and width of a neural
network architecture becomes a key issue in reducing the classification
error. Researchers have been investigating not only neural network architectures
but also the way of utilizing a given dataset. For example, data are augmented
by rotation and translation \cite{cubuk2018autoaugment, inoue2018data,
  takahashi2018ricap}, auxiliary information from external data is fed to a
neural network \cite{torrey2010transfer, lu2015transfer,shin2016deep,
  zamir2018taskonomy}, data are grouped into superclasses in a supervised or
unsupervised way \cite{yan2015hd, chen2019ss}, and information of data is
gradually fed to the neural network \cite{barshan2015stage}.

Note that in conventional use of CNNs for image classification and recognition,
it is assumed that all target classes are equal (i.e., no hierarchy) and
exclusive of one another (i.e., no overlap). CNN-based image classifiers built
on this assumption cannot take into account an innate hierarchy among target
classes (e.g., cats and dogs in animal image classification) or additional
information that can be easily derived from the data (e.g., numbers larger than
five in the recognition of handwritten digits), thereby resulting in scalability
issues when the number of target classes is large.

In this paper, we propose a new learning framework called \textit{hierarchical
  auxiliary learning} based on two related but slightly different ideas of
\textit{hierarchical classification}
\cite{chen2019ss,wu2017hierarchical,yan2015hd,zhu2017b} and \textit{logical
  learning by auxiliary inputs} \cite{wan2018neural}, which not only address the
scalability issues with a large number of classes but also could further reduce
the classification/recognition errors with a reasonable number of classes.
In the hierarchical auxiliary learning, we first group classes into superclasses
(e.g., grouping ``Beagle'' and ``Poodle'' into ``Dog'' and ``Persian Cat'' and
``Russian Blue'' into ``Cat'') and provide this superclass information to a
neural network based on the following three steps: First, the neural network is
augmented with the auxiliary block which takes superclass information and
generates an auxiliary score. We use ResNet \cite{2015arXiv151203385H} as an
example neural network architecture in this paper and insert the auxiliary block
between the last residual block and the fully connected output layer. Second, a
superclass is semantically or non-semantically assigned to each image and
one-hot encoded. Finally, the one-hot-encoded superclass vector is fed to the
auxiliary block and the multiplication of the output of the last residual block
by the output of the auxiliary block is injected to the fully-connected output
layer.

The rest of the paper is organized as follows: Section~\ref{sec:work} introduces
work related with the hierarchical auxiliary learning. Section~\ref{sec:hal}
describes the neural network architecture based on the proposed hierarchical
auxiliary learning. Section~\ref{sec:experimental-results} presents experimental
results which show the classification performance improved by the hierarchical
auxiliary learning and the effect of different superclasses on the
performance. Section~\ref{sec:concluding-remarks} concludes our work in this
paper.

\section{Related Work}
\label{sec:work}
It is well known that transferring learned information to a new task as an
auxiliary information enables efficient learning of a new task
\cite{pan2009survey}, while providing acquired information from a wider network
to a thinner network improves the performance of the thinner network
\cite{romero2014fitnets}.

Auxiliary information from the input data also improves the performance. In the
stage-wise learning, coarse to finer images, which are subsampled from the
original images, are fed to the network step by step to enhance the learning
process \cite{barshan2015stage}. The ROCK architecture introduces an auxiliary
block which can perform multiple tasks of extracting useful information from the
input and inserting it to the input for a main task \cite{mordan2018revisiting}.

There have been proposed numerous approaches to utilize hierarchical class
information as well. \citet{cerri2014hierarchical} connect multi-layer
perceptrons (MLPs) and let each MLP sequentially learn a hierarchical class as
rear layer takes the output of the preceding layer as its
input. \citet{yan2015hd} insert coarse category component and fine category
component after a shared layer. Classes are classified into K-coarse categories,
and K-fine category components are targeted at each coarse category. In
\cite{chen2019ss}, CNN learns label generated by maximum margin clustering at
root node, and images in the same cluster are classified at leaf node.

B-CNN learns from coarse features to fine features by calculating loss between
superclasses and outputs from the branches of the architecture \cite{zhu2017b},
where the loss of B-CNN is the weighted sum of all losses over branches. In
\cite{wu2017hierarchical}, an ultrametric tree is proposed based on semantic
meaning of all classes to use hierarchical class information. The probability of
each node of the ultrametric tree is the sum of the probabilities of leaves
(which has a path from the leaves to the node) and all nodes on the path from
the leaves to the node.

Furthermore, auxiliary inputs are used to check logical reasoning in
\cite{wan2018neural}. Auxiliary inputs based on human knowledge are provided to
the network to let the network learn logical reasoning. The network verifies the
logical information with the auxiliary inputs first and proceeds to the next
stage.

\begin{figure}
  \centering \includegraphics[width=\linewidth]{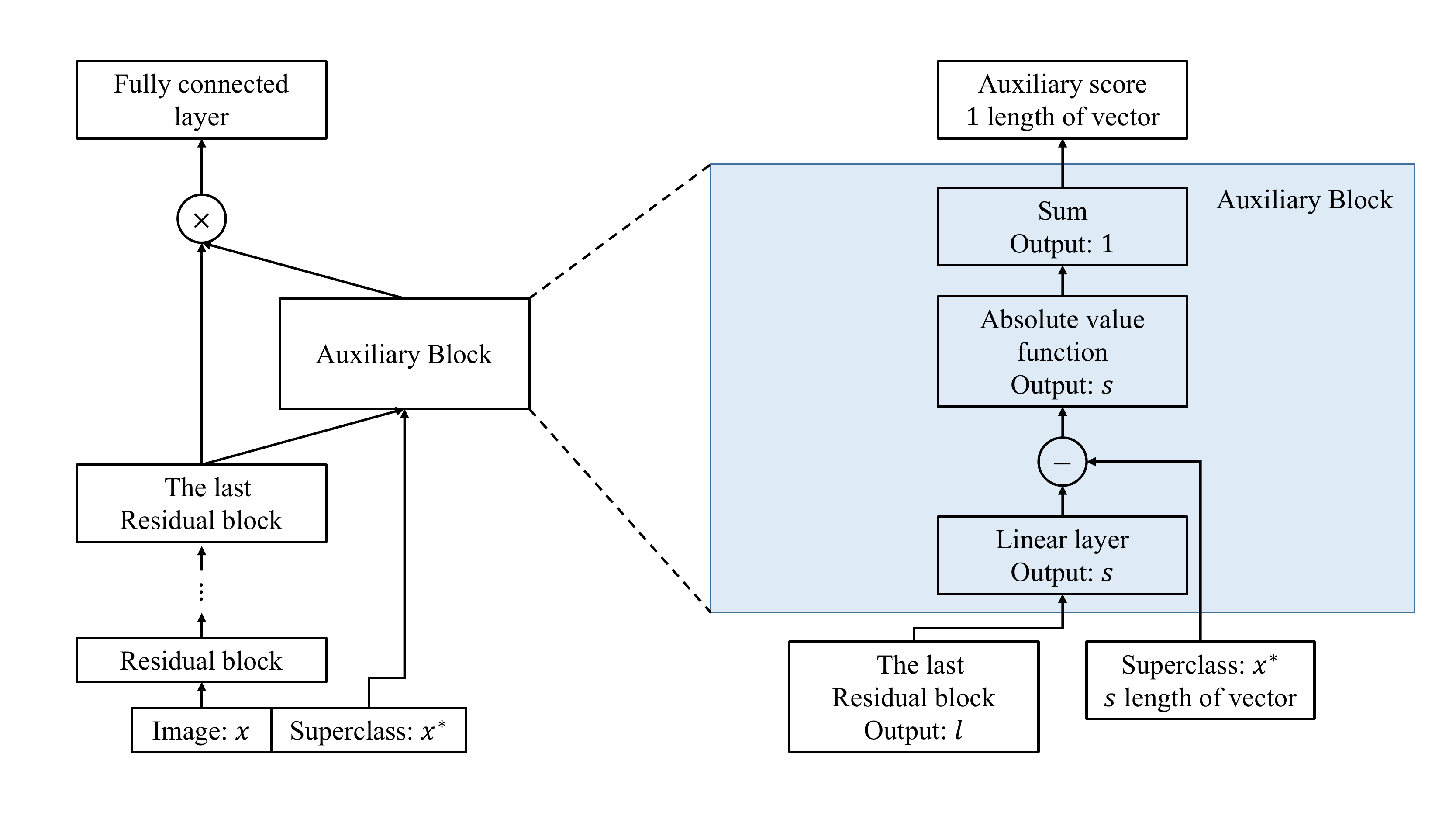}
  \caption{Neural network architecture for the hierarchical auxiliary learning.}
  \label{fig:archi}
\end{figure}

\section{Hierarchical Auxiliary Learning}
\label{sec:hal}
Learning step by step makes it efficient and easy. Most learning forms
hierarchical structure. In the case of image classification learning, images can
be classified through several steps. For example, digits from 0 to 9 can be
grouped into two groups based on the condition that the digit is greater than or
equal to 5. Another example is a dataset consisting of mammal, birds, fish, car,
airplane, and electrical devices. The dataset can be grouped into two
superclasses according to its aliveness. If images are hierarchically
classified, the task becomes easier and more efficient, especially when the
large number of data and classes are present. The goal of the hierarchical
auxiliary learning is to utilize superclasses: For example, digits from 0 to 4
and from 5 to 9 can be grouped into superclass 0 and 1, respectively. Let
$(\bm{x}, \bm{y})$ be a pair of an image and a class. The superclass
$\bm{x^{*}}$ is given to the pair. Hence each element of the dataset now
consists of 3 components, $(\bm{x}, \bm{x^{*}}, \bm{y})$. Unlike the
conventional neural networks, $\bm{x}$ and $\bm{x^{*}}$ is injected to the
neural network. Therefore, the goal becomes to learn a function $f$,
\begin{equation}
  \bm{y_d} = f(\bm{x}, \bm{x^{*}}),
\end{equation}
which minimizes the loss between $y_d$ and $y$.

\subsection{Learning Scheme}
In order to take the advantage of superclass, we introduce an auxiliary
block. It takes superclass information of inputs and utilize it to improve the
performance of the neural network. The auxiliary block can be located between
any two consecutive layers. In this paper, a small ResNet is used as a baseline
and the auxiliary block is located between the last residual block and the
fully-connected output layer as shown in Figure~\ref{fig:archi}. With
one-hot-encoded superclass vector, forward pass is as follows: First, the
auxiliary block takes the superclass vector whose size is batch${\times}s$ and
the output of the last residual block whose size is batch${\times}l$. Then,
batch${\times}l$ vector passes linear layer, and auxiliary score whose size is
batch${\times}1$ is obtained by element-wise subtraction and summation over each
row as described in Figure~\ref{fig:archi}. Finally, the output of the last
residual block is element-wise multiplied by the auxiliary score.

\subsection{Backpropagation}
\label{subsec:back}
The weights of the neural network are adjusted to reduce the error of loss
function through the backpropagation \cite{rumelhart1988learning}. The key point
of the auxiliary block is that the auxiliary score is not directly calculated
from the superclass but learned through the backpropagation. Let $\bm{y^{N-1}}$,
$\bm{y^{N-2}}$, and $a$ be the input to the fully-connected output layer, the
output of the last residual block and the auxiliary score,
respectively. Compared to the original ResNet, the input to the output layer
$\bm{y^{N-1}}$ in the proposed architecture is calculated by
\begin{equation}
  \bm{y^{N-1}} = \bm{y^{N-2}} \times a
\end{equation}
and
\begin{equation}
  a = \sum_j \left|\sum_i w_{ij}\cdot y_i^{N-2} - x^{*}_j\right| ,
\end{equation}
where $w_{ij}$'s are weights of the linear layer in the auxiliary block and
$\bm{x^{*}} = (x^{*}_j)$ is the one-hot-encoded superclass vector. Then, the
neural network is trained as follows:
\begin{equation}
  \frac{\partial L}{\partial a} = \sum_i \frac{\partial L}{\partial y_i^{N-1}}\cdot y_i^{N-2},
\end{equation}
\begin{equation}
  \frac{\partial L}{\partial y_i^{N-2}} = \frac{\partial L}{\partial y_i^{N-1}}\cdot a + \sum_j \chi_j \cdot \frac{\partial L}{\partial a} \cdot w_{ij}
\end{equation}
and
\begin{equation}
  \frac{\partial L}{\partial w_{ij}} = \chi_j \cdot \frac{\partial L}{\partial
    a} \cdot y_i^{N-2} ,
\end{equation}
where
\begin{equation}
  \chi_j = \left\{
    \begin{array}{l l}
      1 & \quad \text{if \, $\sum_i w_{ij}\cdot y_i^{N-2} > x^{*}_j$},\\
      -1 & \quad \text{otherwise}
    \end{array} \right.
\end{equation}
in backward pass where $L$ is a loss function of the neural network.

\section{Experimental Results}
\label{sec:experimental-results}
We evaluate the proposed model on three standard benchmark datasets:
MNIST\footnote{Available at http://www.cs.nyu.edu/~roweis/data.html}, SVHN and
CIFAR-10. The residual block in \cite{he2016identity} is used for baseline in
our experiment.
A 10-layer ResNet is trained on MNIST data and 28-layers ResNet is trained on
SVHN and CIFAR-10. Cosine annealing \cite{loshchilov2016sgdr} is used for
learning rate with maximum learning rate of 1.0 and the minimum learning rate of
0. Batch size and epoch are set to 128 and 250, respectively, for all
datasets. Weights are initialized by He initialization \cite{he2015delving}. All
datasets are cropped after adding 4 additional pixels on each side, randomly
flipped along with horizontal axis and normalized on training. Only
normalization is applied to all datasets during a test stage.

Let $c$ be the number of superclasses. From the label $0$ to $c{-}1$ becomes
$[1,0,{\cdots},0]$ to $[0,{\cdots},0,1]$ in order to be fed to the auxiliary
block. Therefore, $x^{*}_j$ in Section~\ref{subsec:back} takes $1$ if its
superclass is $j$. Otherwise, it is $0$.

\begin{table}
  \caption{Superclass information}
  \label{tbl:super}
  \centering
  \begin{tabular}{l|l|ll}
    \toprule
    Dataset & Case &  Semantics & Superclass \\
    \midrule
    MNIST    & case1 & $\geq$ 5    & 0:$\{$5, 6, 7, 8, 9$\}$, 1:$\{$0, 1 ,2 ,3, 4$\}$ \\
    SVHN     & case2 & mod 2     & 0:$\{$1, 3, 5, 7, 9$\}$, 1:$\{$0, 2, 4, 6, 8$\}$ \\
            & case3 & prime & 0:$\{$2, 3, 5, 7$\}$, 1:$\{$0, 1, 4, 6, 8, 9$\}$\\ 
            & case4 & circle/ curve/ straight line & 0:$\{$0, 6, 8, 9$\}$, 1:$\{$2, 3, 5$\}$, 2:$\{$1, 4, 7$\}$ \\

    \midrule
    
    CIFAR-10    & case1 &   None  & 0:$\{$5, 6, 7, 8, 9$\}$, 1:$\{$0, 1 ,2 ,3, 4$\}$ \\
            & case2 & None     & 0:$\{$1, 3, 5, 7, 9$\}$, 1:$\{$0, 2, 4, 6, 8$\}$ \\
            & case3 & transportation/ animal & 0:$\{$2, 3, 4, ,5, 6, 7$\}$, 1:$\{$0, 1, 8, 9$\}$\\
            & case4 & car/ small animal/ big animal/  & 0:$\{$1, 9$\}$, 1:$\{$3, 5$\}$, 2:$\{$4, 7$\}$,  \\
            & & craft/ others & 3:$\{$0, 8$\}$, 4:$\{$2, 6$\}$\\

    \bottomrule
  \end{tabular}
\end{table}
\begin{table}
  \caption{Error comparison}
  \label{tbl:error}
  \centering
  \begin{tabular}{l|lllll}
    \toprule
    MNIST    & baseline     & case1 & case2 &  case3 & case4\\
    \midrule
    Error & 0.93  & 0.43 $\pm$ 0.03 & 0.73 $\pm$ 0.06 &  0.70 $\pm$ 0.05 & 0.69 $\pm$ 0.00    \\
    \bottomrule  
  
    \toprule
    SVHN    & baseline     & case1 & case2 &  case3 & case4\\
    \midrule
    Error & 4.05  & 2.53 $\pm$ 0.06 & 2.64 $\pm$ 0.11  & 2.66 $\pm$ 0.07 & 2.86  $\pm$ 0.07  \\
    \bottomrule

    \toprule
    CIFAR-10    & baseline     & case1 & case2 &  case3 & case4\\
    \midrule
    Error & 6.81  & 3.30 $\pm$ 0.06 & 5.30 $\pm$ 0.14 & 6.46 $\pm$ 0.08 & 5.13 $\pm$ 0.09    \\
    \bottomrule
  \end{tabular}
\end{table}

\subsection{MNIST}
\begin{figure}
  \begin{minipage}{0.24\linewidth}
    \centering
    \includegraphics[width=\linewidth]{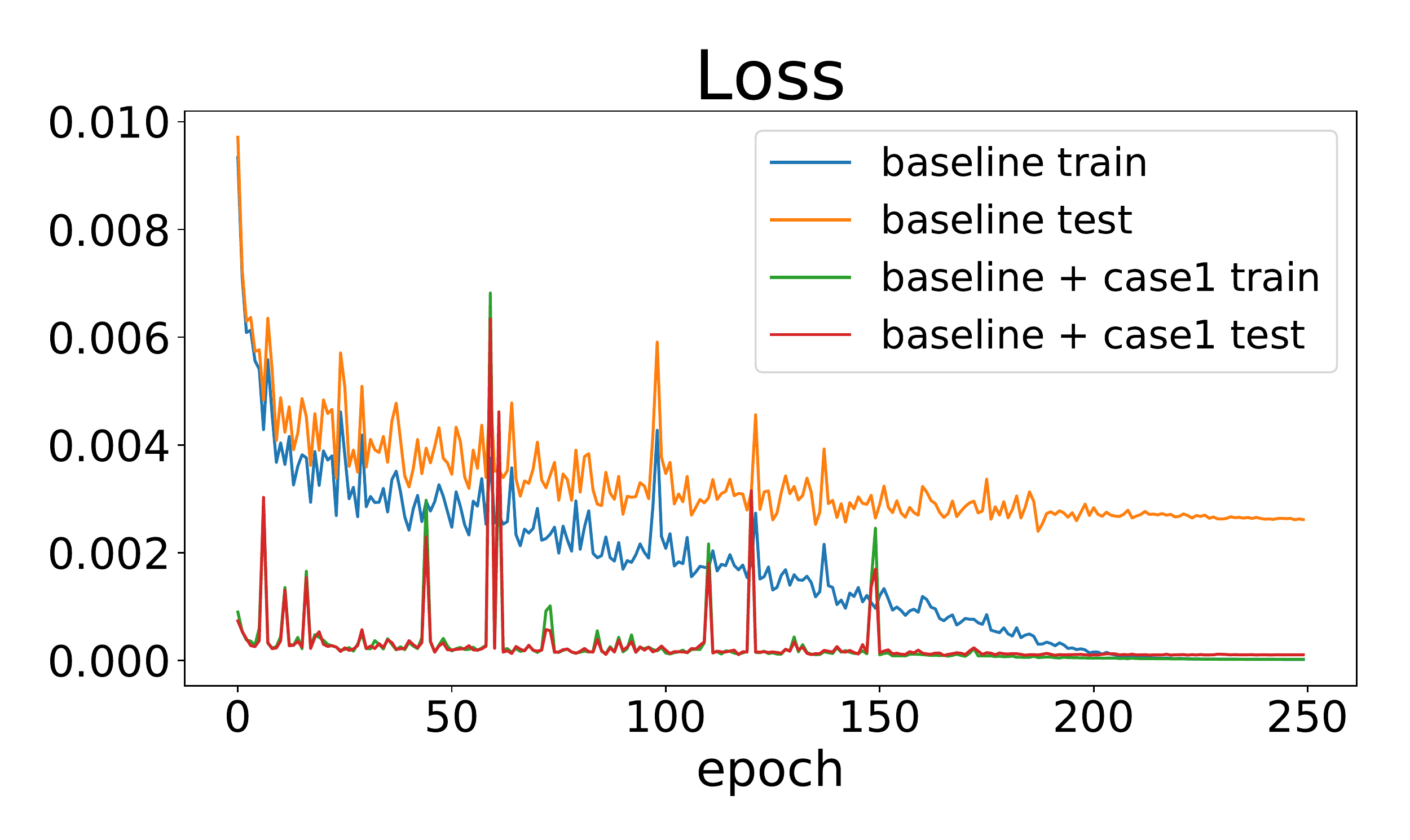}\\
    {\scriptsize (a)}
  \end{minipage}
  \hfill
  \begin{minipage}{0.24\linewidth}
    \centering
    \includegraphics[width=\linewidth]{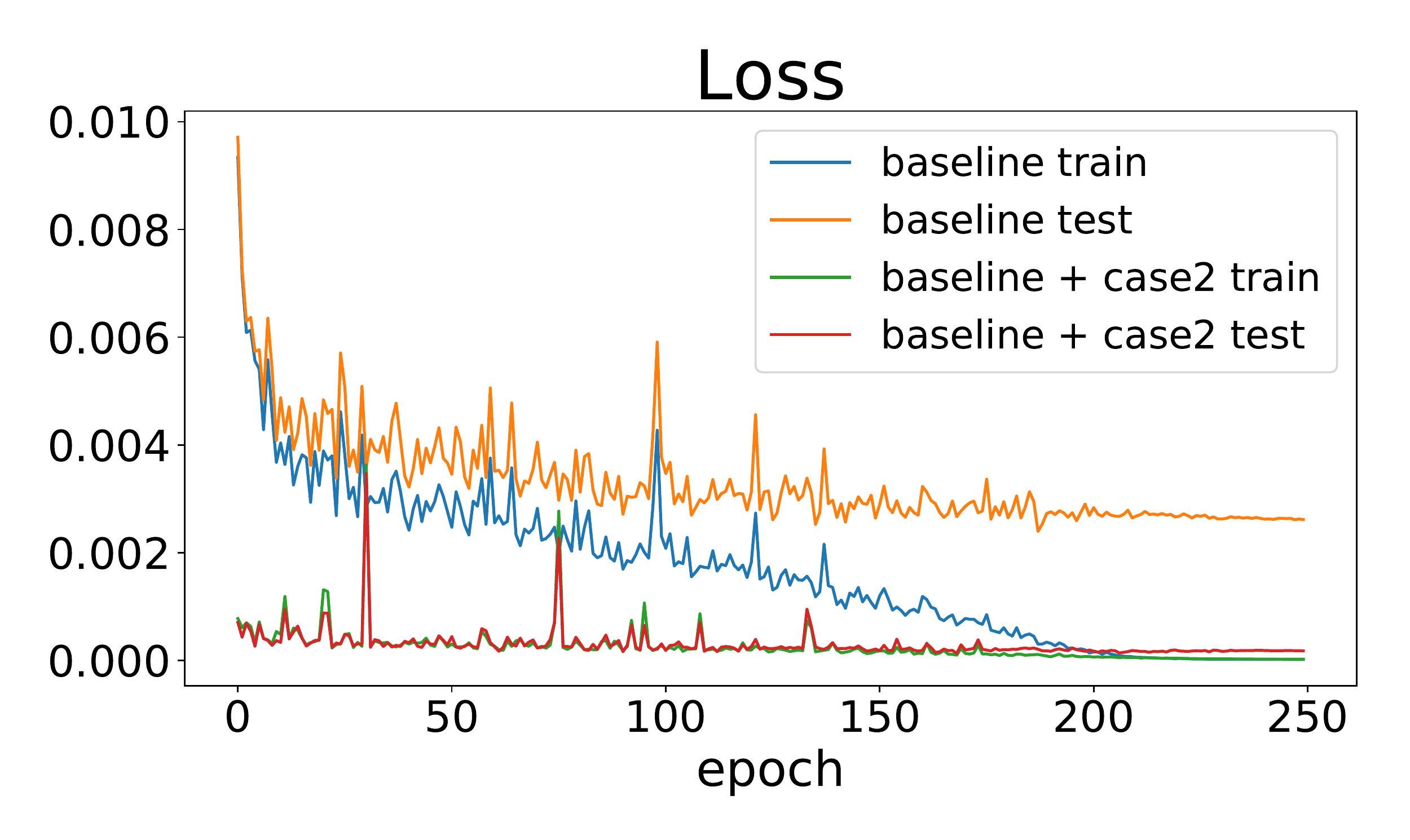}\\
    {\scriptsize (b)}
  \end{minipage}
  \hfill
  \begin{minipage}{0.24\linewidth}
    \centering
    \includegraphics[width=\linewidth]{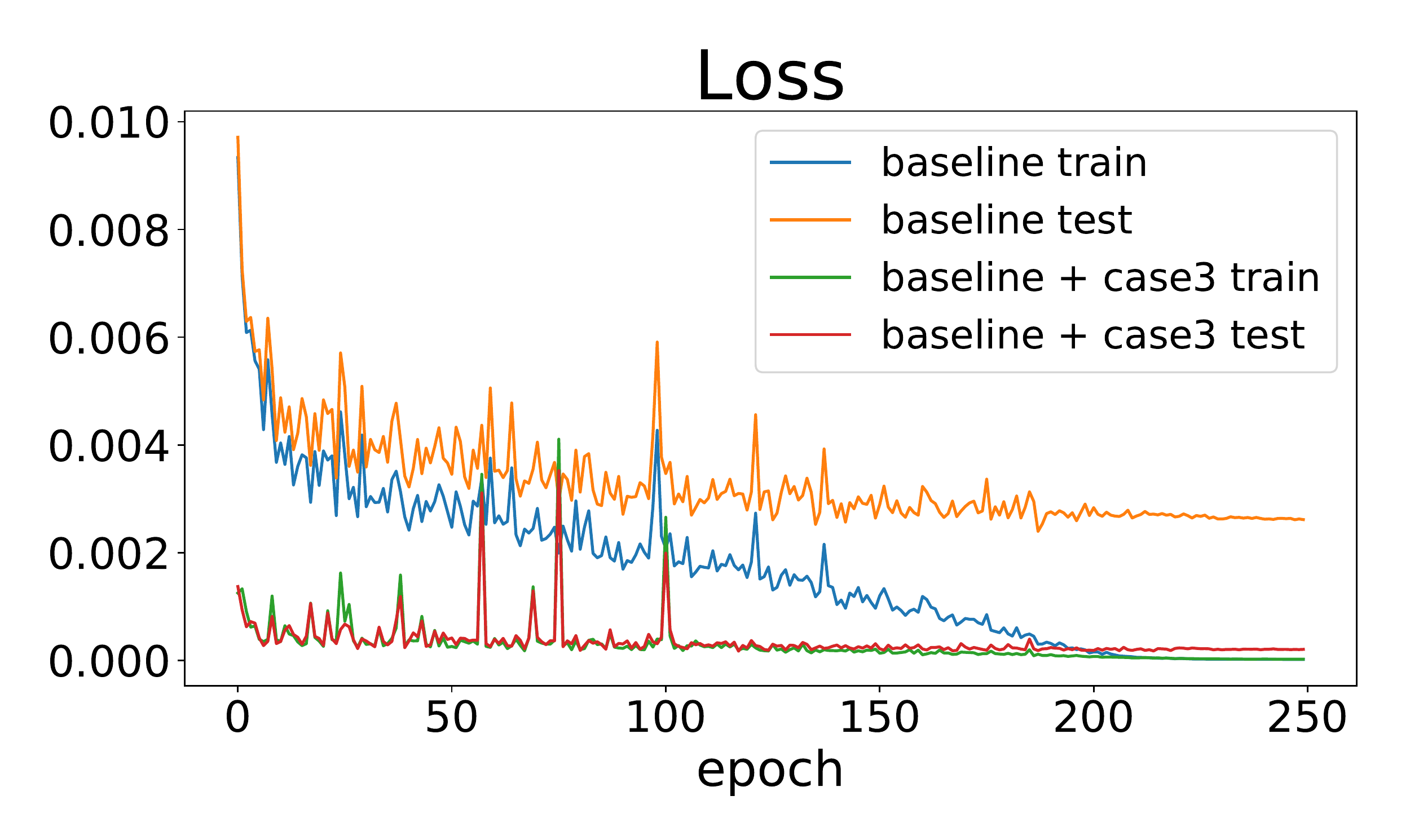}\\
    {\scriptsize (c)}
  \end{minipage}
  \hfill
  \begin{minipage}{0.24\linewidth}
    \centering
    \includegraphics[width=\linewidth]{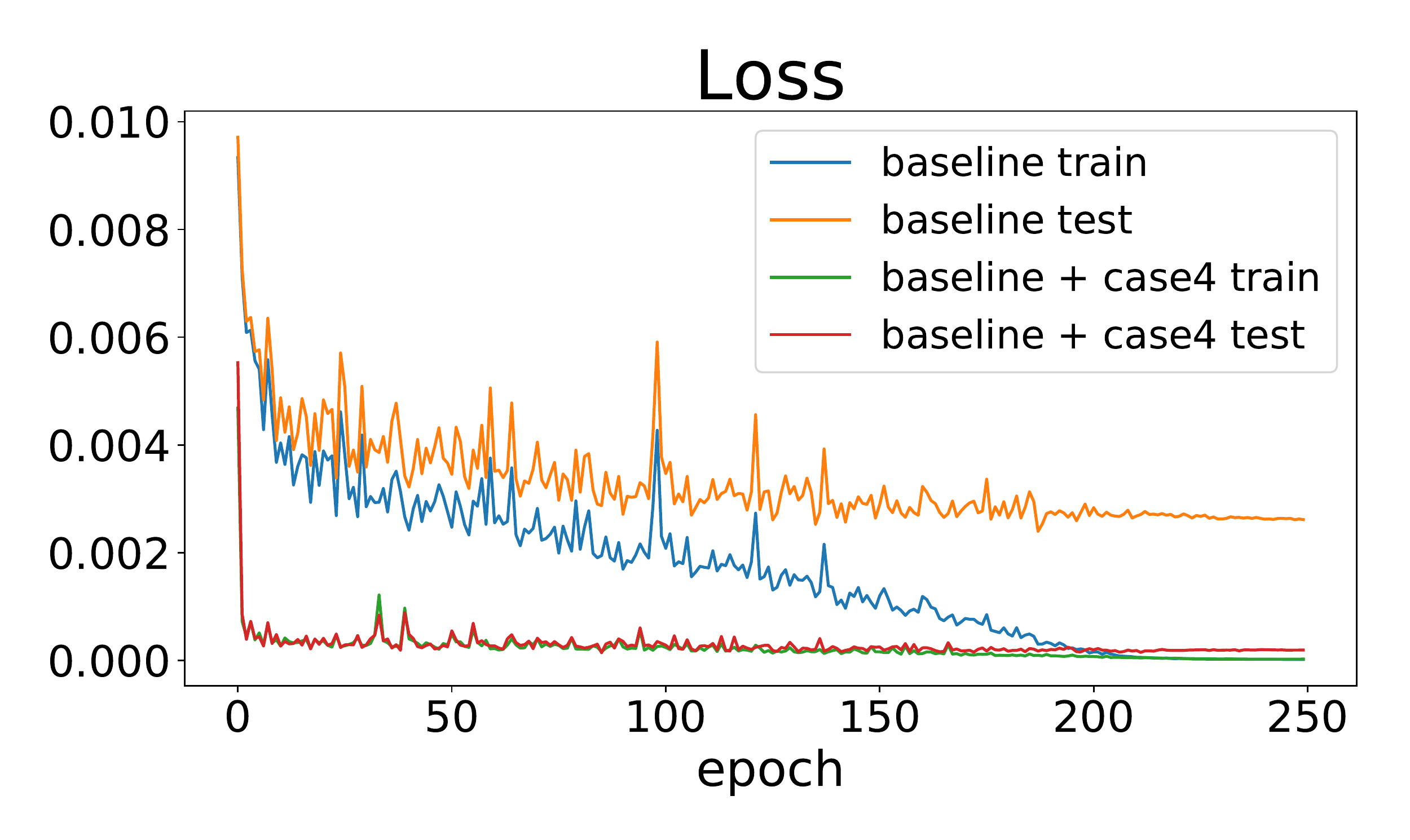}\\
    {\scriptsize (d)}
  \end{minipage}
  \caption{Loss comparison of train and test dataset at each epoch during
    training between the baseline and (a) case1 (b) case2 (c) case3 and (d)
    case4 with MNIST dataset.}
  \label{fig:mnist_loss}
  \begin{minipage}{0.24\linewidth}
    \centering
    \includegraphics[width=\linewidth]{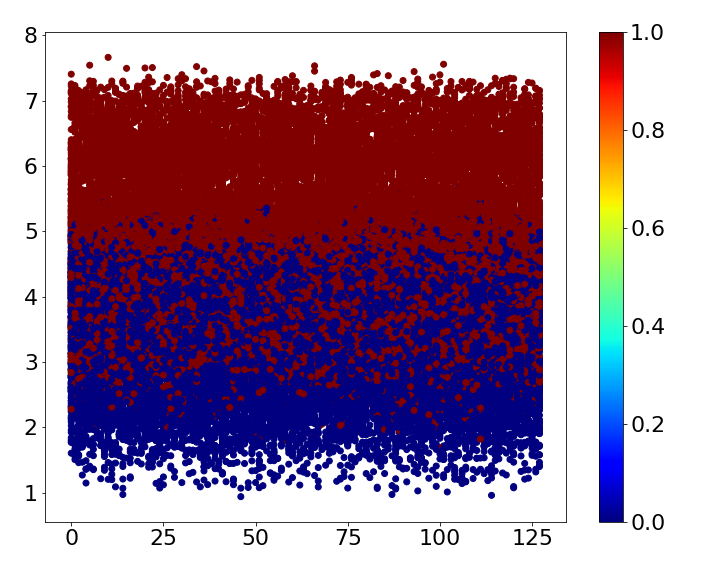}\\
    {\scriptsize (a)}
  \end{minipage}
  \hfill
  \begin{minipage}{0.24\linewidth}
    \centering
    \includegraphics[width=\linewidth]{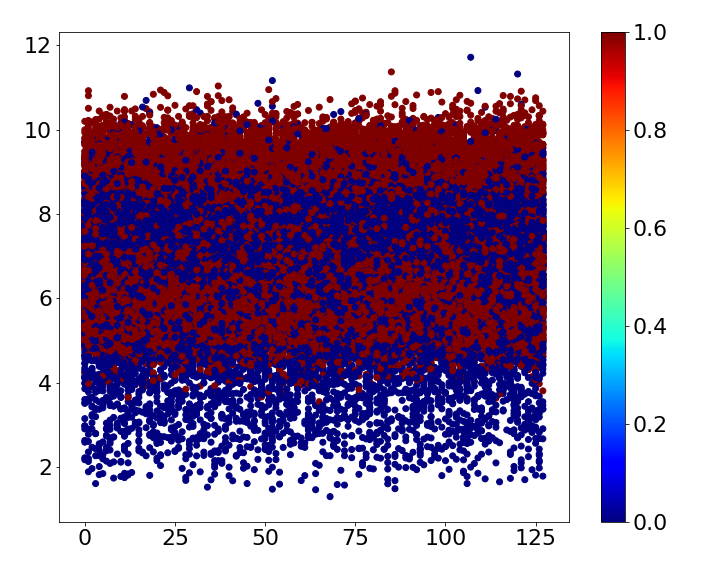}\\
    {\scriptsize (b)}
  \end{minipage}
  \hfill
  \begin{minipage}{0.24\linewidth}
    \centering
    \includegraphics[width=\linewidth]{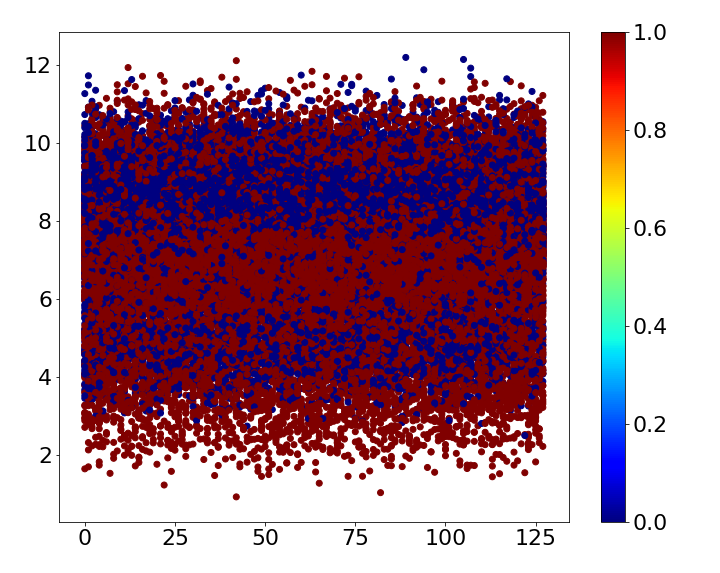}\\
    {\scriptsize (c)}
  \end{minipage}
  \hfill
  \begin{minipage}{0.24\linewidth}
    \centering
    \includegraphics[width=\linewidth]{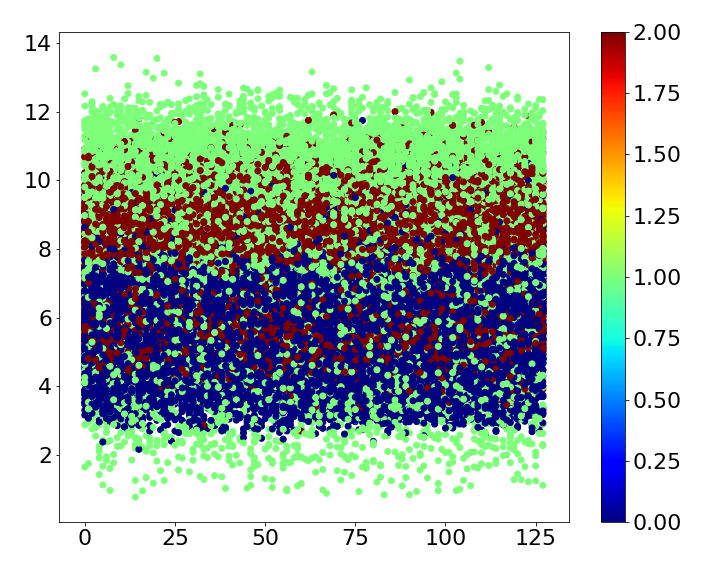}\\
    {\scriptsize (d)}
  \end{minipage}
  \caption{Auxiliary scores of all training images corresponding to their
    superclass of (a) case1, (b) case2, (c) case3, and (d) case4 with MNIST
    dataset.
  }
  \label{fig:mnist_aux_score}
\end{figure}

MNIST is one of the most widely used benchmark datasets in classification. It is
composed of 10 classes of handwritten digits form $0$ to $9$. The size of each
image is $28{\times}28$ and the number of training and test images are 60,000
and 10,000, respectively. Superclass of the dataset is given by three different
ways based on human knowledge and one way based on the shape of digit. First,
the dataset is divided into two superclasses by five. Second, we split the
dataset into even and odd numbers. Third, prime number forms the same
superclass. Finally, superclass is determined by noticing the shape of digits:
Digit $0, 6, 8$ and $9$ have a circle shape, and digit $2, 3$ and $5$ do not
have a circle shape but have a curve. The rest of digits has straight lines
only. The four different ways of assigning superclasses are summarized in
Table~\ref{tbl:super}. All cases are trained five times, and the mean and
standard deviation are shown in Table~\ref{tbl:error}. While the baseline
mismatches $0.93\%$, all cases mentioned above reduce the error irrespective of
whether the superclass is given based on human knowledge or image itself. Loss
of train and test dataset while training the baseline and each case is shown in
Figure~\ref{fig:mnist_loss}. Due to the auxiliary score, the loss of train and
test dataset for all the cases shows faster convergence than the
baseline. Figure~\ref{fig:mnist_aux_score} shows the auxiliary scores of all
training images obtained after 250 epochs of training. The auxiliary scores of
case1, which has the lowest error, show clear separation between the auxiliary
scores corresponding to the two superclasses. Otherwise, the auxiliary scores of
superclasses are mixed for all other cases, which result in lower error
reductions than case1.

\subsection{SVHN}
\begin{figure}
  \begin{minipage}{0.24\linewidth}
    \centering
    \includegraphics[width=\linewidth]{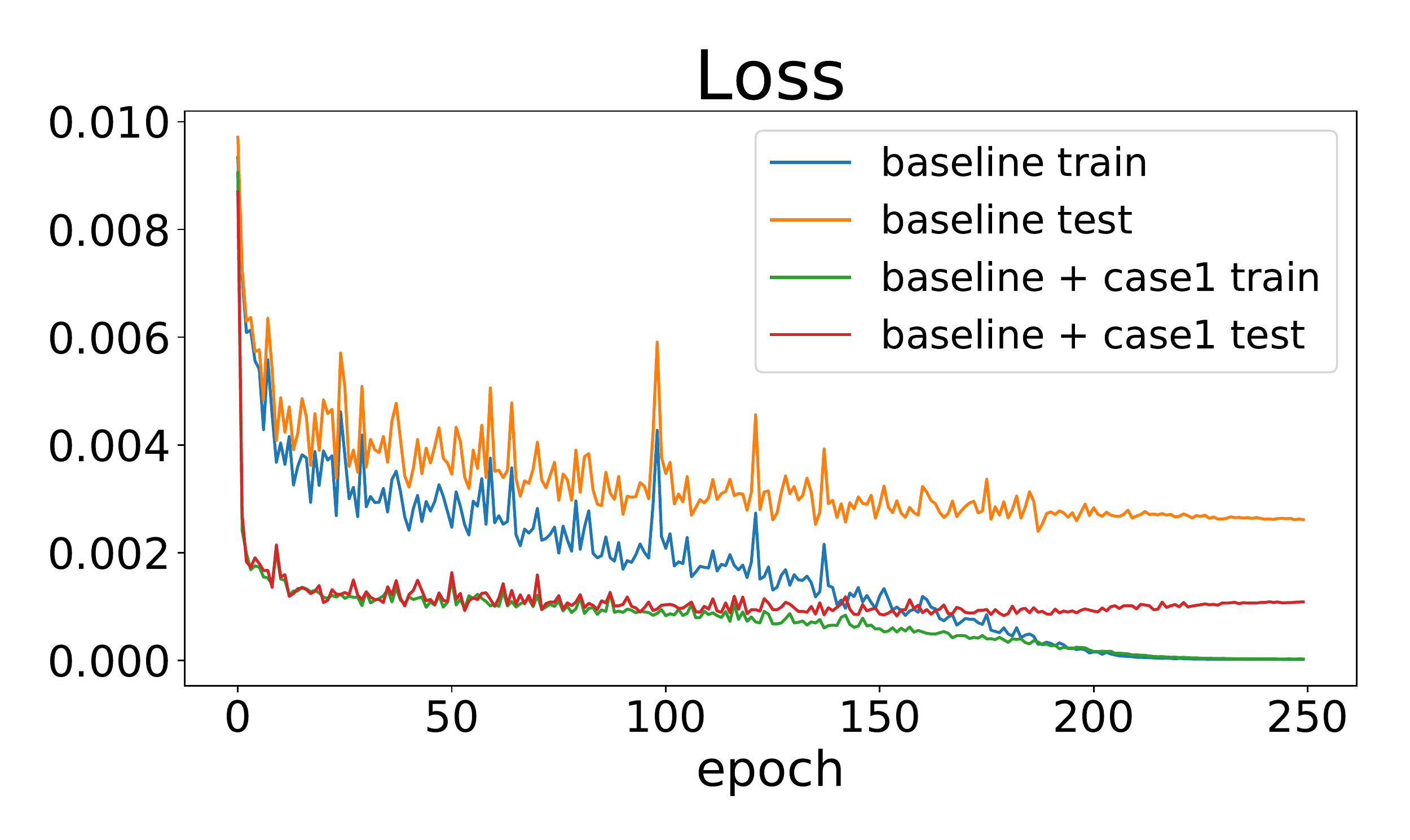}\\
    {\scriptsize (a)}
  \end{minipage}
  \hfill
  \begin{minipage}{0.24\linewidth}
    \centering
    \includegraphics[width=\linewidth]{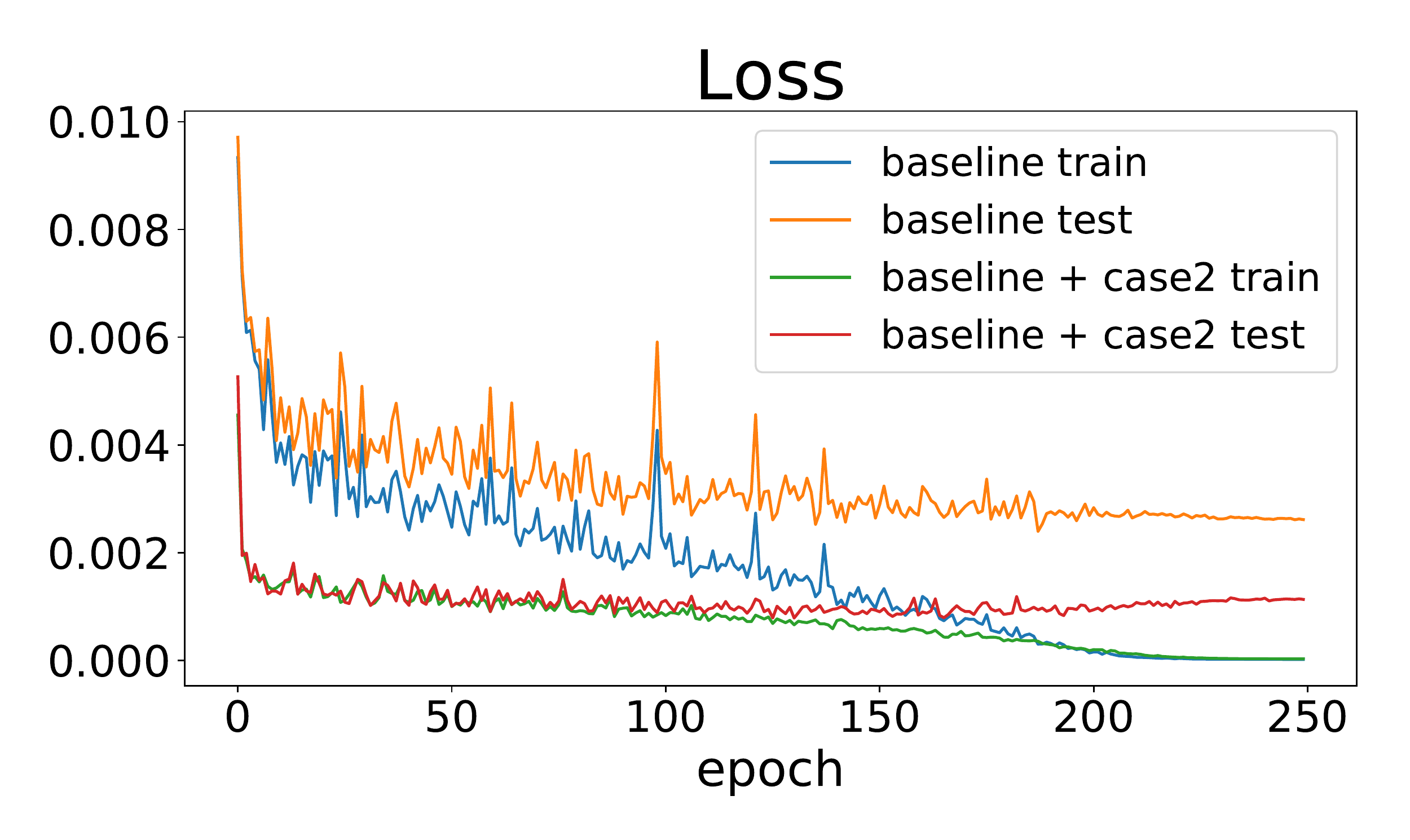}\\
    {\scriptsize (b)}
  \end{minipage}
  \hfill
  \begin{minipage}{0.24\linewidth}
    \centering
    \includegraphics[width=\linewidth]{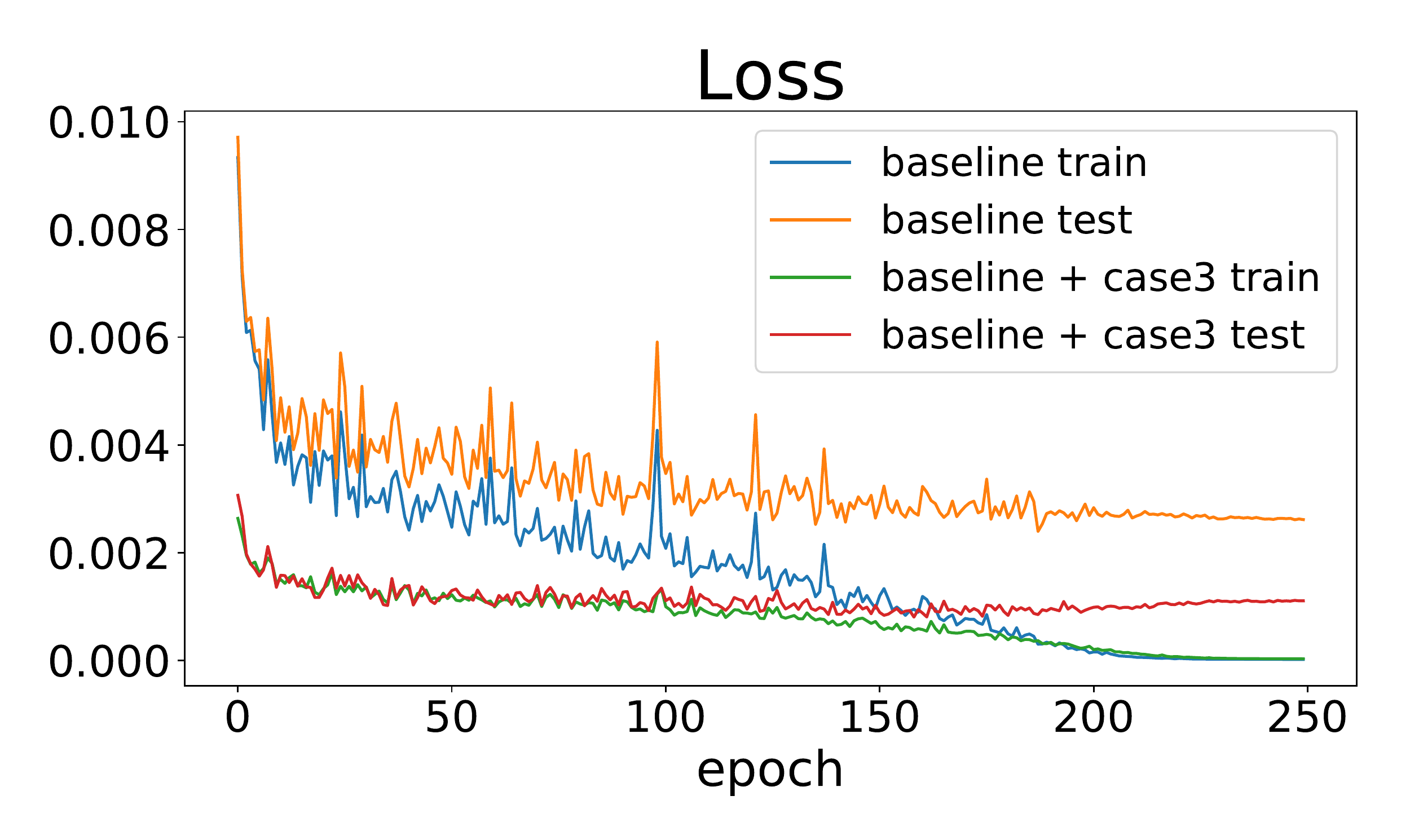}\\
    {\scriptsize (c)}
  \end{minipage}
  \hfill
  \begin{minipage}{0.24\linewidth}
    \centering
    \includegraphics[width=\linewidth]{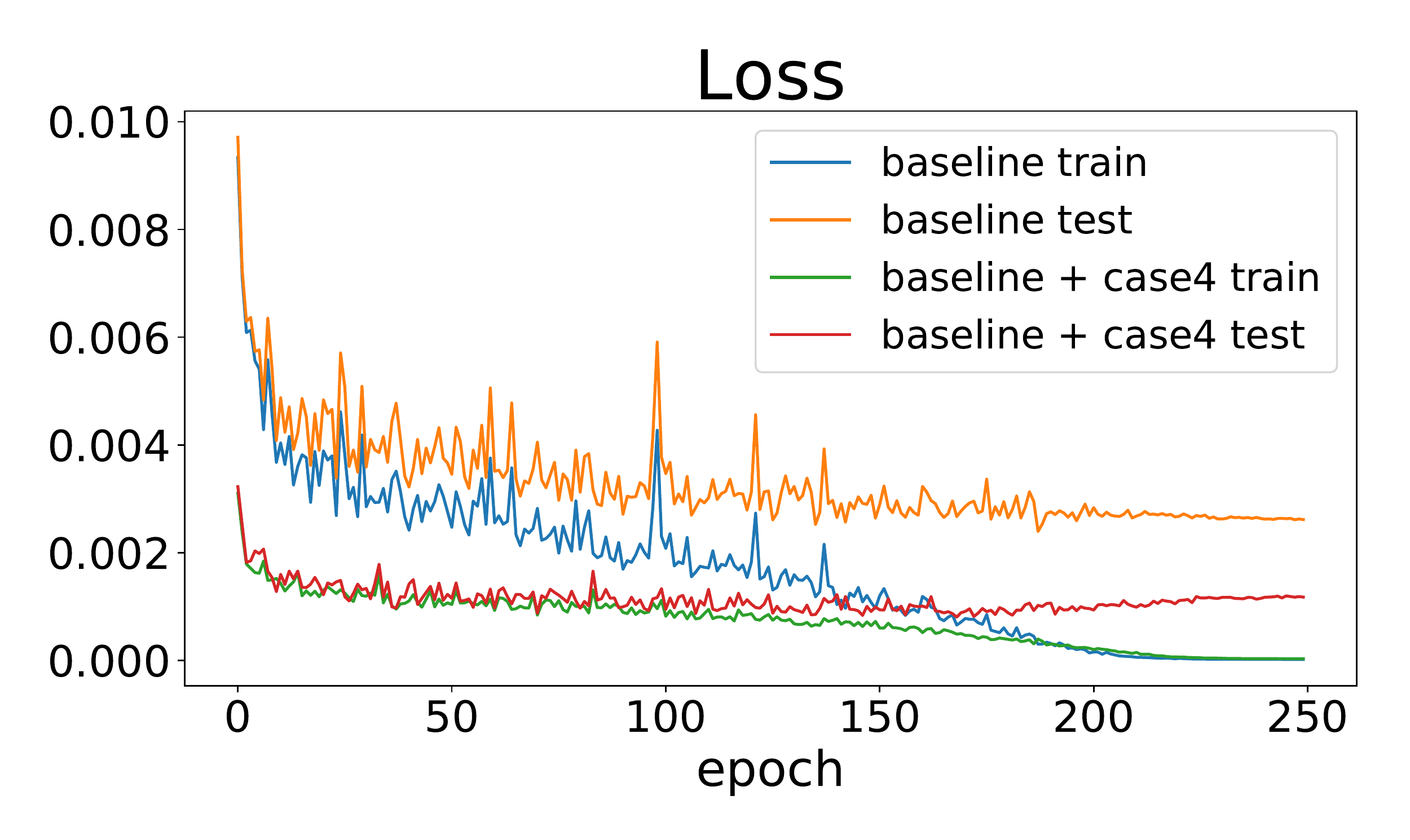}\\
    {\scriptsize (d)}
  \end{minipage}
  \caption{Loss comparison of train and test dataset at each epoch during
    training between the baseline and (a) case1 (b) case2 (c) case3 and (d)
    case4 with SVHN dataset.}
  \label{fig:svhn_loss}
  \begin{minipage}{0.24\linewidth}
    \centering
    \includegraphics[width=\linewidth]{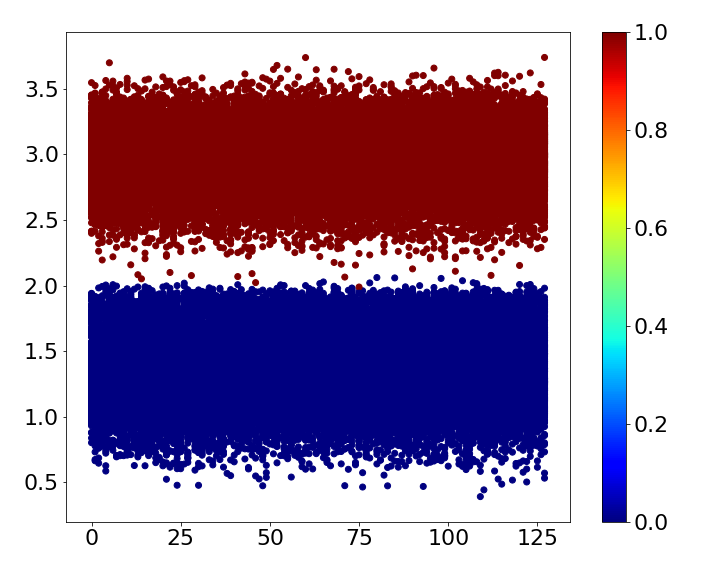}\\
    {\scriptsize (a)}
  \end{minipage}
  \hfill
  \begin{minipage}{0.24\linewidth}
    \centering
    \includegraphics[width=\linewidth]{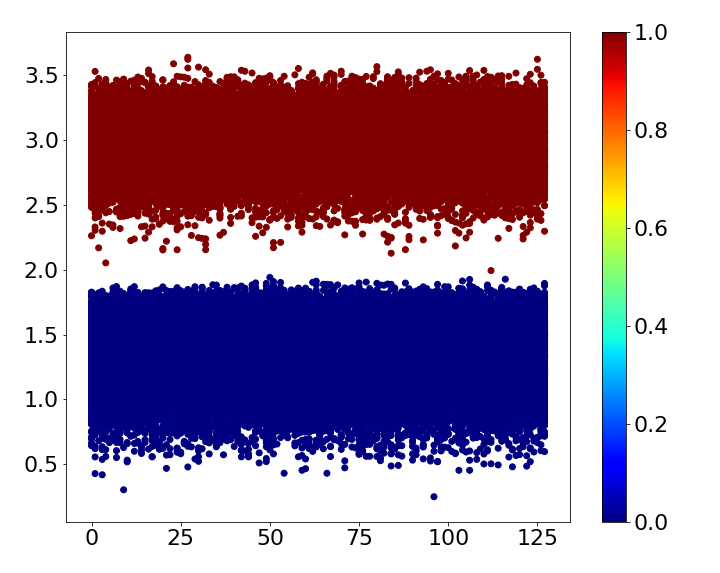}\\
    {\scriptsize (b)}
  \end{minipage}
  \hfill
  \begin{minipage}{0.24\linewidth}
    \centering
    \includegraphics[width=\linewidth]{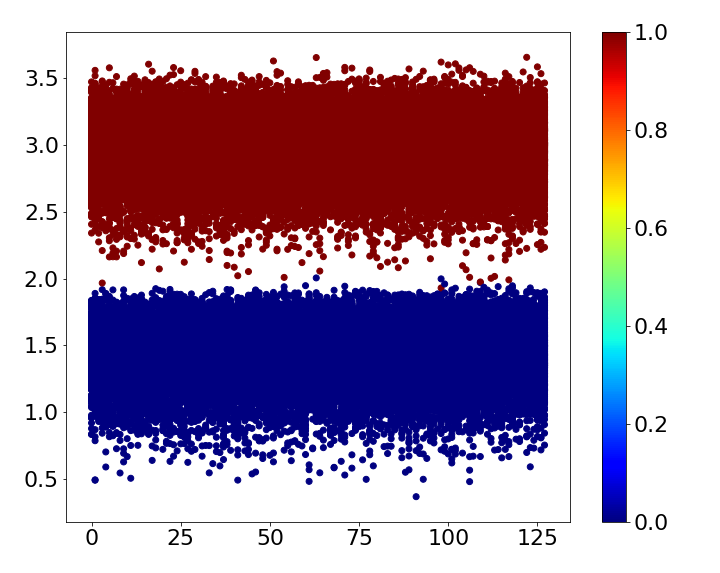}\\
    {\scriptsize (c)}
  \end{minipage}
  \hfill
  \begin{minipage}{0.24\linewidth}
    \centering
    \includegraphics[width=\linewidth]{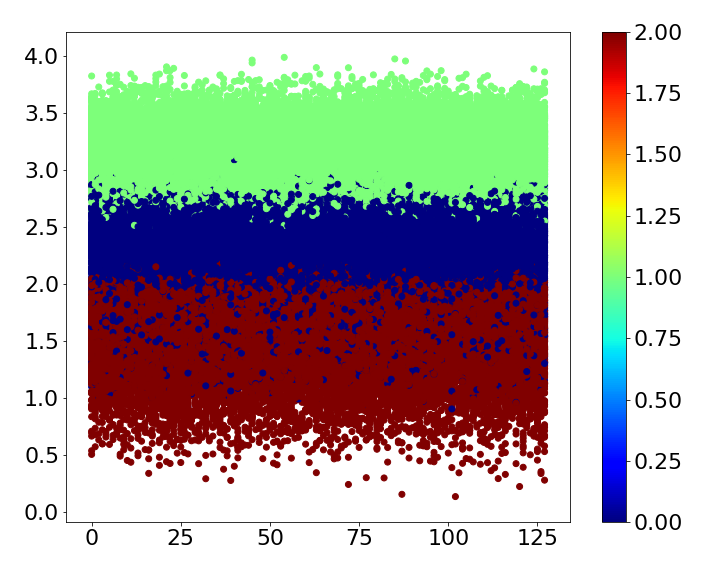}\\
    {\scriptsize (d)}
  \end{minipage}
  \caption{Auxiliary scores of all training images corresponding to their
    superclass of (a) case1, (b) case2, (c) case3, and (d) case4 with SVHN
    dataset.
  }
  \label{fig:svhn_aux_score}
\end{figure}

The Street View House Numbers (SVHN) data set also has 10 digits like
MNIST. While MNIST is handwritten digit, SVHN consists of digits from a real
world with the size of $32{\times}32$ \cite{netzer2011reading}. 73,257 training
images and 26,032 test images are available in the dataset. Because it has the
same class with the MNIST dataset, we train SVHN with the four cases used for
MNIST. Each case is trained 5 times and the mean of errors is shown in
Table~\ref{tbl:error}. The results demonstrate that all cases improve the
accuracy at least $1.2\%$. As observed with MNIST dataset,
Figure~\ref{fig:svhn_loss} shows that the loss of train and test dataset shows
faster convergence when superclasses are introduced to the network. The
auxiliary scores of all cases are well split according to their superclass as
shown in Figure~\ref{fig:svhn_aux_score}. As a result, differences between error
reduction of case1, case2 and case3 are not significant. In addition, case4,
which has 3 superclasses, also results in higher than $1\%$ error reduction as
its auxiliary scores are well divided into 3 layers according to their
superclasses.
\begin{table}
  \caption{CIFAR-10 classes}
  \label{tbl:cifar10}
  \centering
  \begin{tabular}{c|cccccccccc}
    \toprule
    Name & airplane  & car & bird  &  cat & deer & dog & frog & horse & ship & truck \\
    \midrule
    Label    & 0 & 1 & 2 & 3 & 4 & 5 & 6 & 7 & 8 & 9   \\
    \bottomrule  
  \end{tabular}
\end{table}

\subsection{CIFAR-10}
\begin{figure}
  \begin{minipage}{0.24\linewidth}
    \centering
    \includegraphics[width=\linewidth]{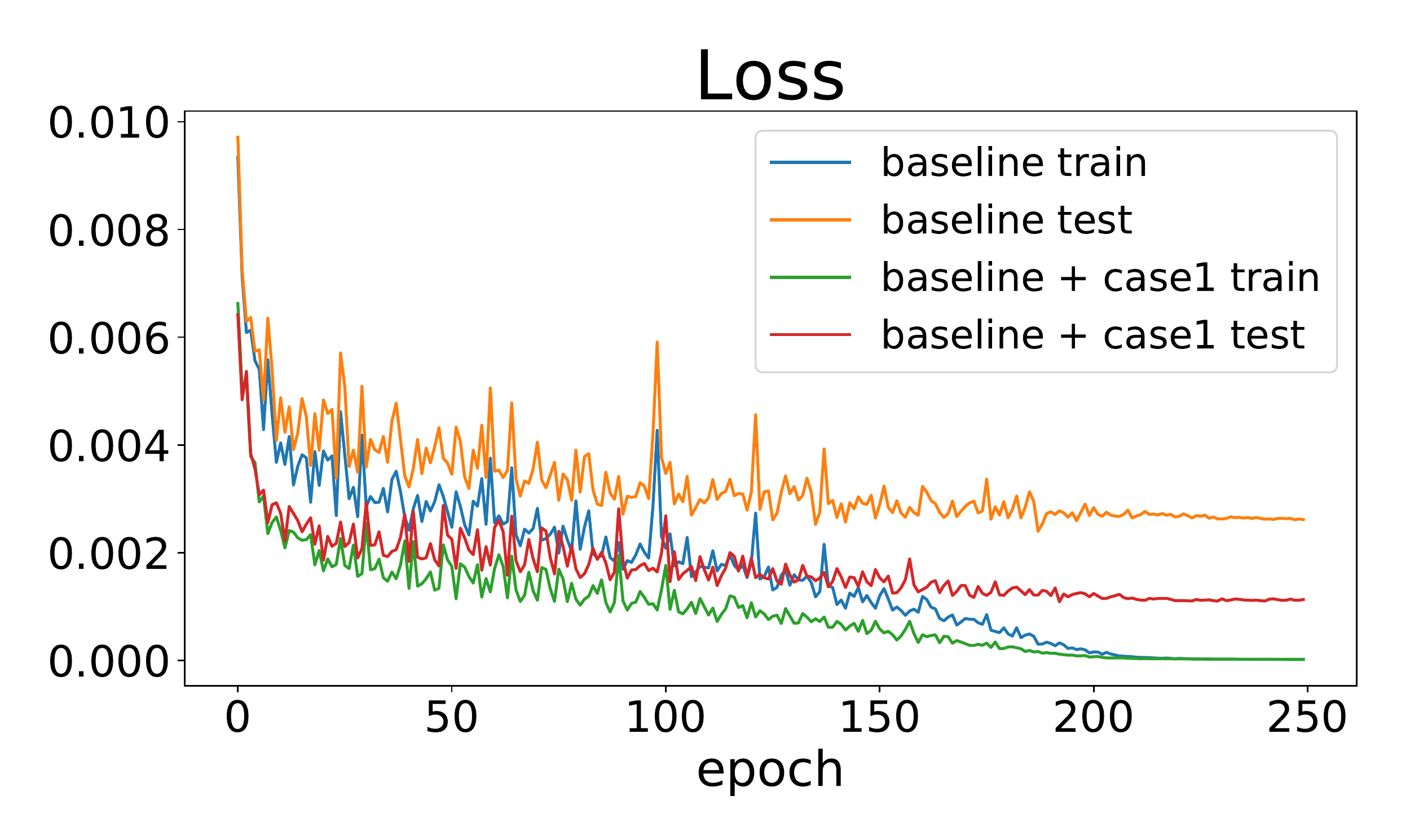}\\
    {\scriptsize (a)}
  \end{minipage}
  \hfill
  \begin{minipage}{0.24\linewidth}
    \centering
    \includegraphics[width=\linewidth]{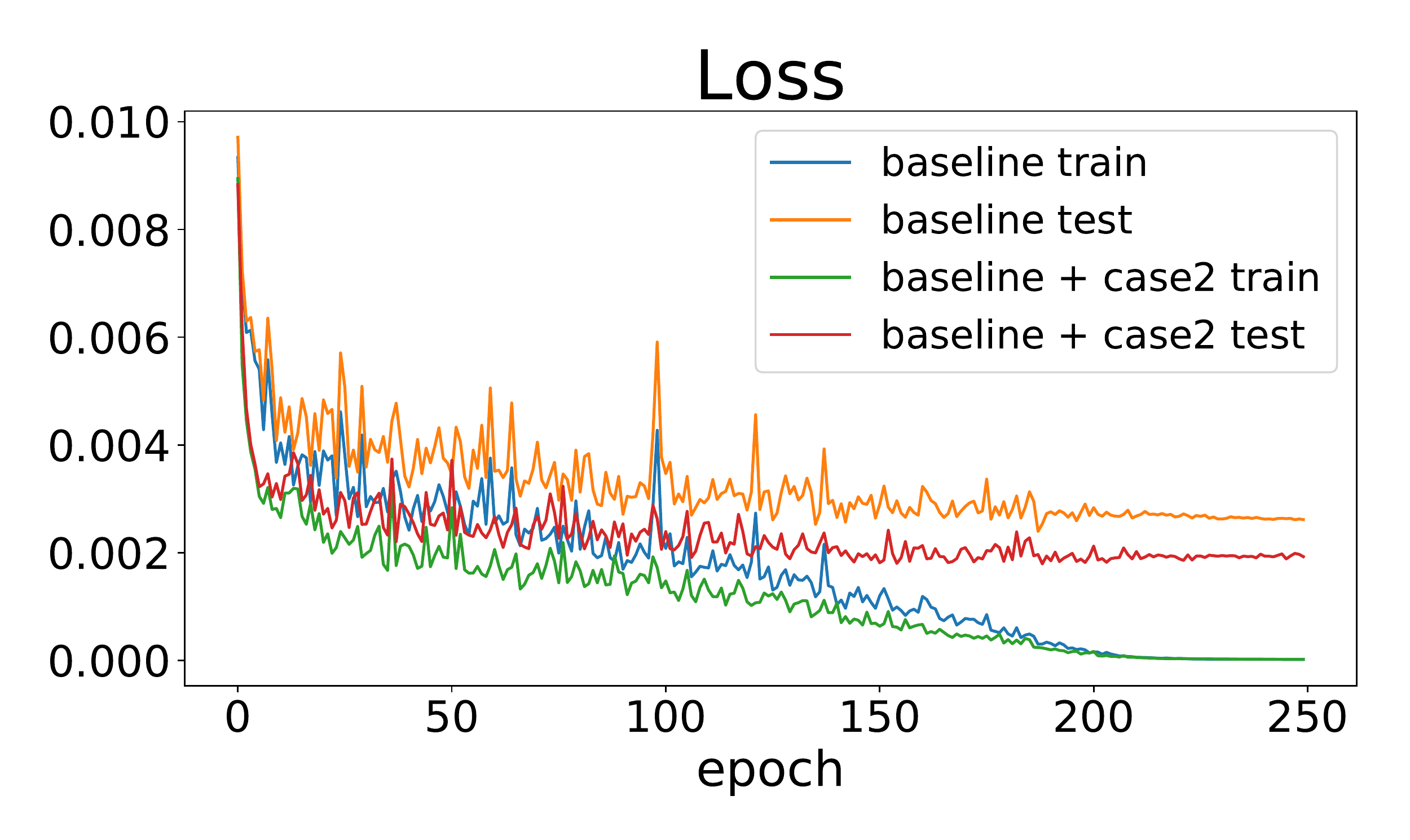}\\
    {\scriptsize (b)}
  \end{minipage}
  \hfill
  \begin{minipage}{0.24\linewidth}
    \centering
    \includegraphics[width=\linewidth]{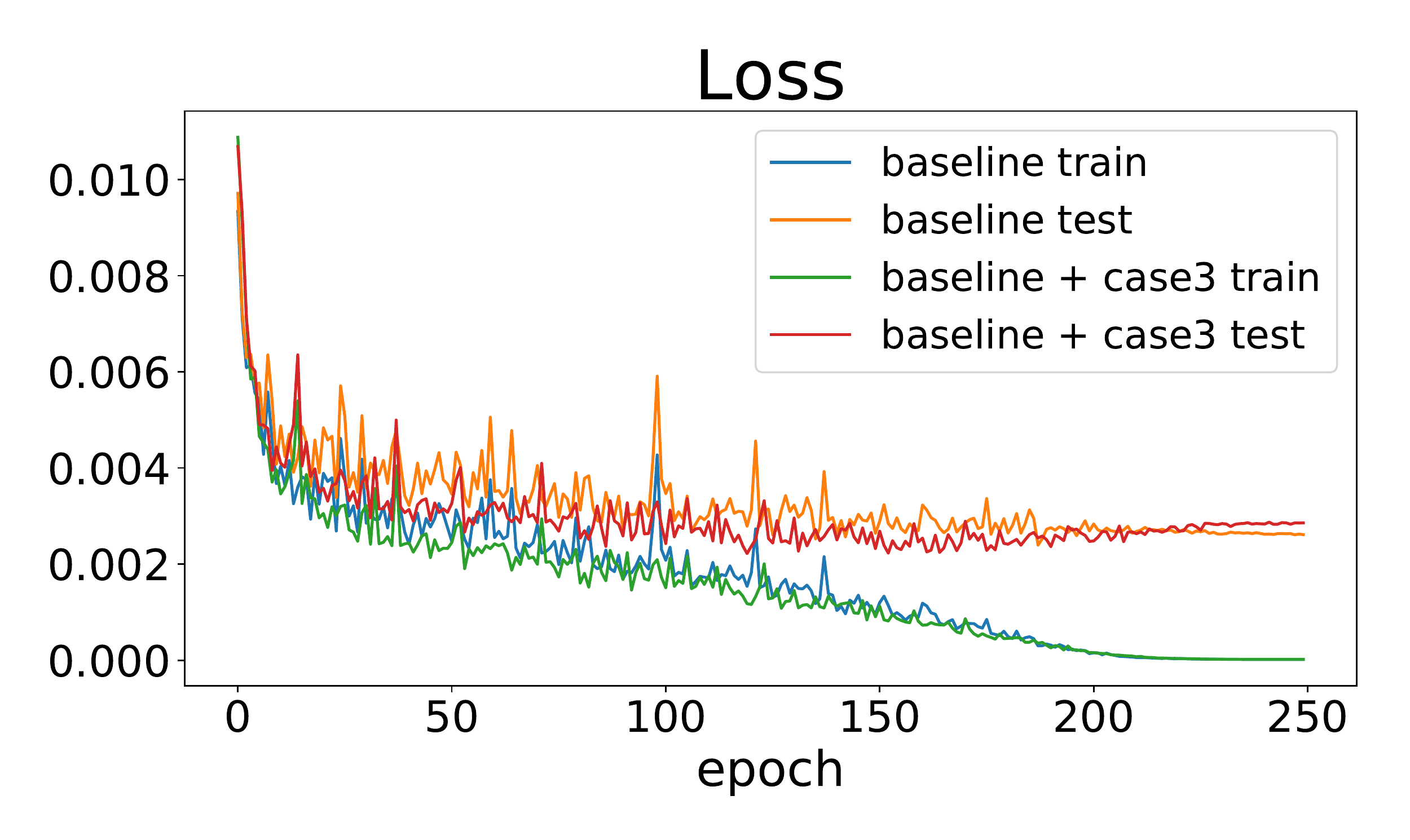}\\
    {\scriptsize (c)}
  \end{minipage}
  \hfill
  \begin{minipage}{0.24\linewidth}
    \centering
    \includegraphics[width=\linewidth]{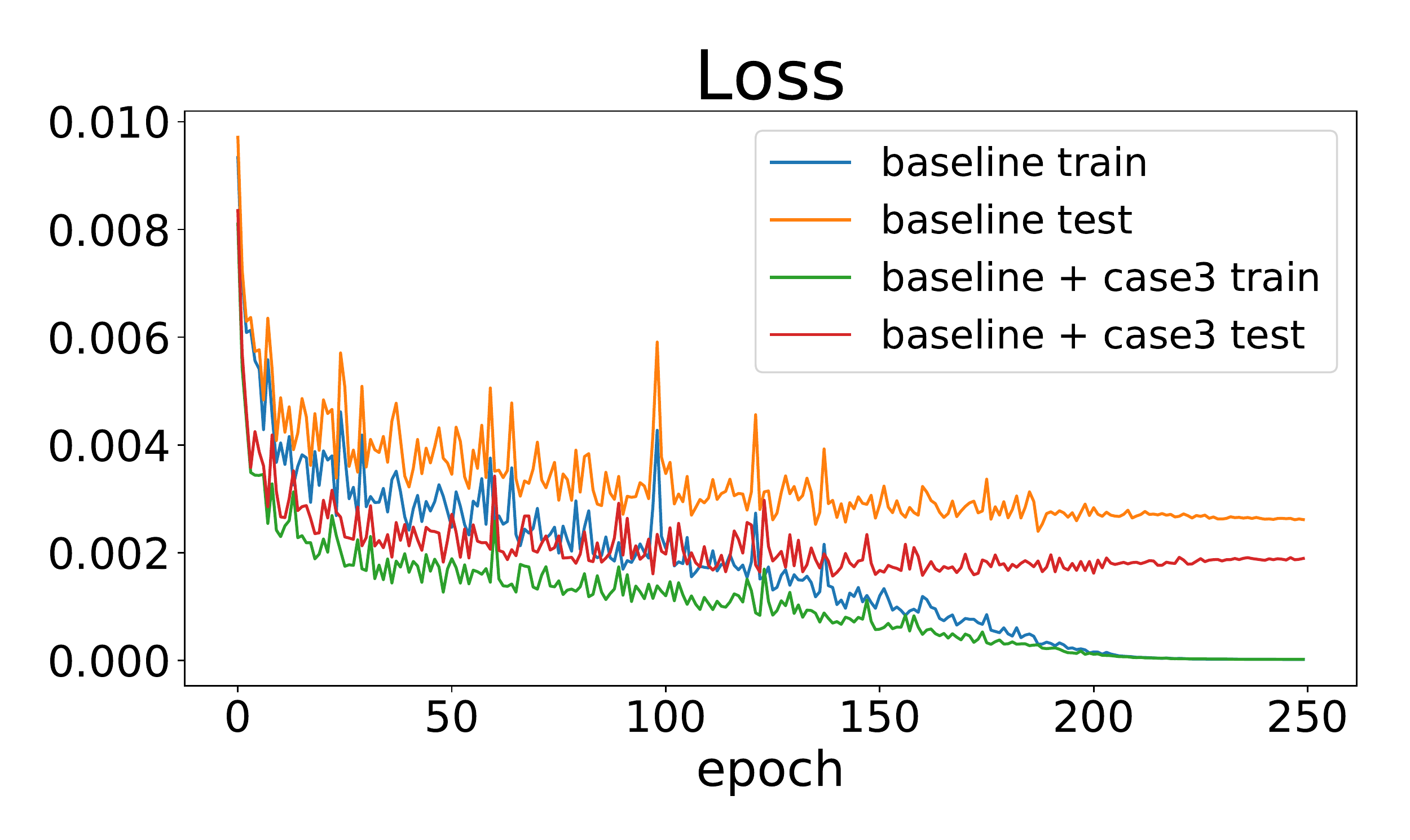}\\
    {\scriptsize (d)}
  \end{minipage}
  \caption{Loss comparison of train and test dataset at each epoch during
    training between the baseline and (a) case1 (b) case2 (c) case3 and (d)
    case4 with CIFAR-10 dataset.}
  \label{fig:cifar10_loss}
  \begin{minipage}{0.24\linewidth}
    \centering
    \includegraphics[width=\linewidth]{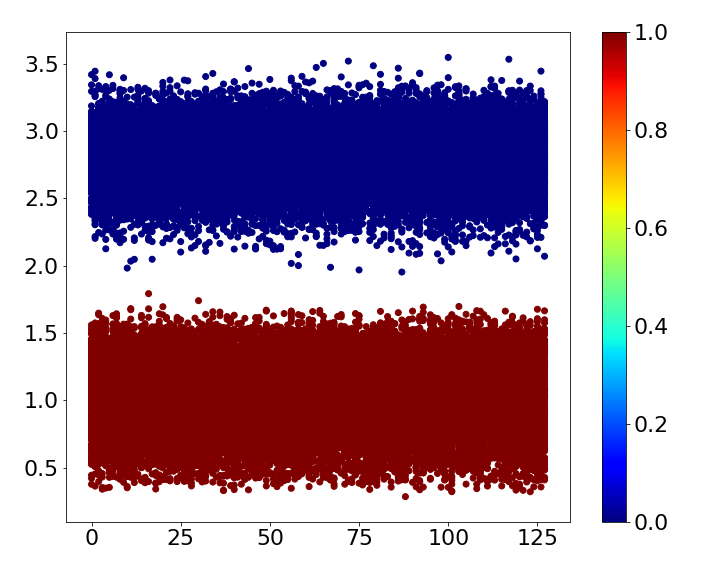}\\
    {\scriptsize (a)}
  \end{minipage}
  \hfill
  \begin{minipage}{0.24\linewidth}
    \centering
    \includegraphics[width=\linewidth]{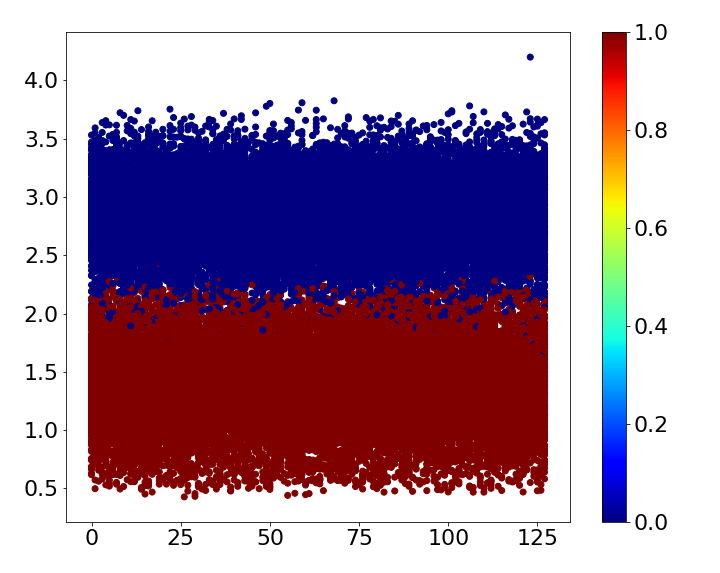}\\
    {\scriptsize (b)}
  \end{minipage}
  \hfill
  \begin{minipage}{0.24\linewidth}
    \centering
    \includegraphics[width=\linewidth]{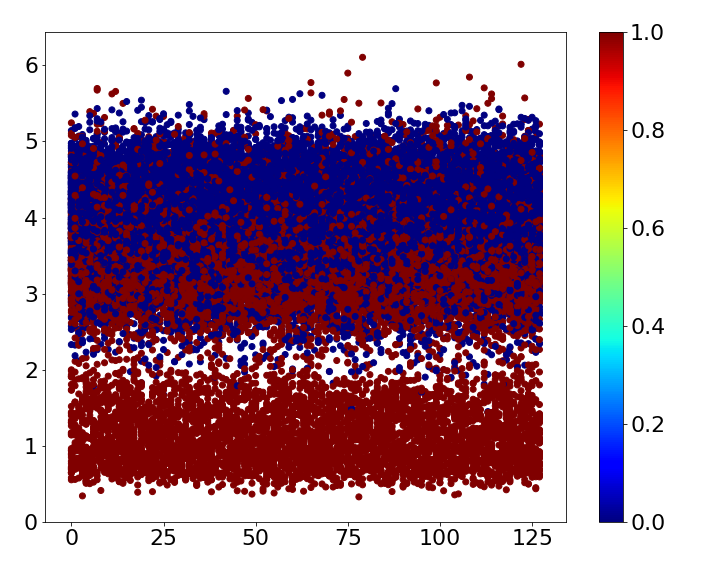}\\
    {\scriptsize (c)}
  \end{minipage}
  \hfill
  \begin{minipage}{0.24\linewidth}
    \centering
    \includegraphics[width=\linewidth]{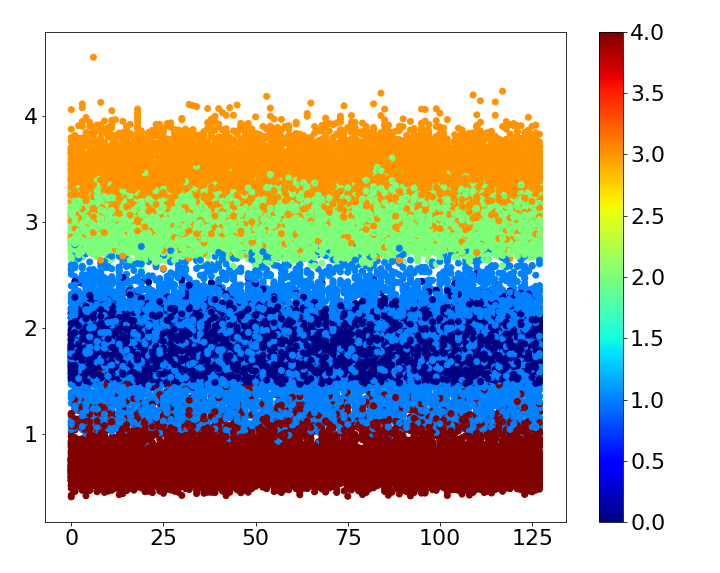}\\
    {\scriptsize (d)}
  \end{minipage}
  \caption{Auxiliary scores of all training images corresponding to their
    superclass of (a) case1, (b) case2, (c) case3, and (d) case4 with CIFAR-10
    dataset.
  }
  \label{fig:cifar10_aux_score}
\end{figure}

CIFAR-10 consists of 60,000 images with the size of $32{\times}32$, which belong
to 10 different classes listed in Table~\ref{tbl:cifar10}. 500 images of each
class form training set and the others are used for a test. We assign
superclasses to images both non-semantically and semantically. First, superclass
is simply given according to its label: If its label is larger than or equal to
5, then superclass 0 is given; otherwise, 1 is given. Second, superclass 0 is
given to a class if a label of the class is an odd number; otherwise, 1 is
given. As shown in Table~\ref{tbl:error}, the non-semantical superclass
assignments improve the performance. Third, classes are semantically grouped
into two superclasses, i.e., transportation and animal. Finally, 5 superclasses
are assigned based on the criteria described in
Table~\ref{tbl:super}. Figure~\ref{fig:cifar10_loss} also shows that learning of
the neural network is faster and more efficient in terms of convergence when
superclasses are used. Split of auxiliary scores shows a stark difference between
case1 and case3 in Figure~\ref{fig:cifar10_aux_score}. Case1---which shows clear
split among auxiliary scores---provides much better classification performance
than case3 as shown in Table~\ref{tbl:error}. Auxiliary scores for case4 which
includes 5 superclasses are divided into almost 5 layers, though the auxiliary
scores of superclasses car and small animals are mixed together.

\section{Concluding Remarks}
\label{sec:concluding-remarks}
To not only address the scalability issues in classification with a large number
of classes but also improve the classification/recognition performance with a
reasonable number of classes, in this paper we have proposed the hierarchical
auxiliary learning, a new learning framework exploiting innate hierarchy among
target classes or additional information easily derived from the data
themselves. Under the proposed learning framework, each image is assigned a
superclass semantically or non-semantically built from classes, which is one-hot
encoded and used to compute an auxiliary score through the auxiliary block. With
the help of the auxiliary score provided by the auxiliary block, the proposed
neural network architecture can improve the performance of the classification in
terms of error and loss. The experimental results demonstrate that the
classification performance of the neural network based on the proposed learning
framework highly depends on how clearly the auxiliary scores are split according
to superclasses, which indicates that further study is required to find a
systematic way of constructing superclasses resulting in clear separation in
auxiliary scores.

\bibliographystyle{plainnat} \bibliography{neurips_2019}

\end{document}